\documentclass{article} 
\usepackage{iclr2020_conference,times}

\usepackage{amsmath,amsfonts,bm}

\def\eqref#1{equation~\ref{#1}}

\def\1{\bm{1}}

\DeclareMathAlphabet{\mathsfit}{\encodingdefault}{\sfdefault}{m}{sl}
\SetMathAlphabet{\mathsfit}{bold}{\encodingdefault}{\sfdefault}{bx}{n}

\usepackage{hyperref}
\usepackage{url}
\usepackage{ifthen}
\usepackage{graphicx}
\usepackage{xcolor}

\title{Exploring Exploration: Comparing Children with RL Agents in Unified Environments}

\author{Eliza Kosoy$^1$\thanks{Corresponding author: \texttt{eko@berkeley.edu}} \hspace{2pt}, Jasmine Collins$^1$, David M. Chan$^1$, Sandy Huang$^2$, Deepak Pathak$^1$,\\
\textbf{Pulkit Agrawal$^3$, John Canny$^1$, Alison Gopnik$^1$, \& Jessica B. Hamrick$^2$}\\
$^1$University of California, Berkeley\\
$^2$DeepMind, London \\
$^3$Massachusetts Institute of Technology \\
}

\iclrfinalcopy 
\begin{document}

\maketitle

\begin{abstract} 

Research in developmental psychology consistently shows that children explore the world thoroughly and efficiently  and that this exploration allows them to learn. In turn, this early learning supports more robust generalization and intelligent behavior later in life.
While much work has gone into developing methods for exploration in machine learning, artificial agents have not yet reached the high standard set by their human counterparts.
In this work we propose using DeepMind Lab \citep{beattie2016deepmind} as a platform to directly compare child and agent behaviors and to develop new exploration techniques.
We outline two ongoing experiments to demonstrate the effectiveness of a direct comparison, and outline a number of open research questions that we believe can be tested using this methodology.
\end{abstract}

\section{The Problem of Exploration}
\label{sec:intro}
The problem of exploration is one of the most fundamental issues in reinforcement learning (RL): how should an agent gather enough experience from different parts of the world in order to later produce approximately optimal behavior?
Questions surrounding exploration have been investigated for as long as the field has existed \citep{thompson1933likelihood,robbins1952some,gittins1979dynamic}, and it continues to be a major focus of research today, with recent approaches estimating various quantities to guide exploration, such as visit counts \citep{bellemare2013arcade,ostrovski2017count,martin2017count,tang2017exploration,machado2018count}, uncertainty \citep{osband2016deep,burda2018exploration}, surprise \citep{schmidhuber1991curious,pathak2017curiosity}, learning progress \citep{kaplan2004maximizing,baranes2013active}, disagreement \citep{pathak2019self}, or other forms of novelty \citep{fu2017ex2}; see \citet{Lavet_2018} for a recent review.
Yet, despite these efforts, the problem of exploration remains far from solved.
Indeed, algorithms which achieve state-of-the-art performance on benchmarks such as Atari often still rely on simple exploration strategies like $\epsilon$-greedy combined with huge amounts of computation \citep{kapturowski2018recurrent}.

As with artificial agents, exploration is a key feature of human behavior.
Dating back to \citet{Piaget1933}, developmental researchers have conceived of children as active and curious learners who are intrinsically motivated to explore the world in systematic and rational ways \citep{Schulz_2007,Cook,Legare_2012,Schulz_2012,Ruggeri_2019}; see \citet{Schulz_2012} for a review.
The simplest example of this exploration may be in the way that active movement through space informs both object understanding and navigation.  
For example, when infants become mobile and begin to crawl this exploration appears to allow them to learn both about space and objects \citep{campos2000travel}. Even 11-month old infants choose to physically explore objects that violate expectations of object solidity or object support instead of a novel distractor object \citep{Stahl_2015}.
Older children also explore in more complex ways, both when evidence does not conform to their expectations \citep{Bonawitz_2012,Legare_2012} and when evidence is causally confounded \citep{Schulz_2007,Glymour_2007,Cook,Schulz_2012}.
The simple fact that children know less than adults may make them more open to new kinds of learning and exploration \citep{Lucas_2014,Gopnik_2017}. Recent evidence suggests that children do indeed explore more than adults, and that this translates into higher amounts of learning, even when exploration comes at a cost \citep{Liquin_2019,schulz2019searching,Sumner_2019,Hartley_2019}.
Moreover, such learning is rapid and supports powerful, abstract generalizations \citep{Xu_2017,Bonawitz_2020}.
For example, \citet{Xu_2017} found that in the course of playing with a toy that was activated by different shaped and colored blocks, preschoolers  could develop an abstract ``overhypothesis'' about how the toy functioned, such as determining whether the blocks worked based on their color or their shape, and use that overhypothesis  to make inferences about a new toy or block. 

The generalization and rapid learning resulting from children's exploration is in stark contrast with that exhibited by modern RL agents.
We suggest that by performing \emph{direct}, controlled comparisons between children and agents, we may be able to leverage insights from children's exploratory behavior to improve the design of RL algorithms.
Although previous approaches to exploration have been motivated qualitatively by human behavior \citep[e.g.][]{pathak2017curiosity}, they typically have not included direct comparisons, making it difficult to know whether such methods actually capture the behavior they are inspired by.
For example, while learning exhibited by children is typically measured over a handful of trials, the learning done by curious RL agents is measured over millions of environment steps.
Moreover, experiments with children have not been performed in the types of navigation-centric environments where agents are often trained; we therefore do not know the extent to which results in the developmental literature even apply to such agents.

Prior work has demonstrated the value of human baselines as a useful comparison for agent behavior in other settings.
For example, the baselines on Atari games provided by \citet{mnih2015human} have been widely influential in RL, providing a point of comparison for hundreds of experiments.
Other work goes beyond using human data as a baseline, for example by using it to illuminate key differences between human and agent priors \citep{dubey2018investigating}.
However, with a few interesting exceptions \citep{bambach2018toddler, seita2019zpd}, most existing comparisons have been done with adults.
We argue instead for using \emph{children} as direct inspiration for research in exploration. Very young children learn extensively, and, unlike adults,  they explore widely, ubiquitously and effectively with little direct training, explicit education or reflection.  In fact, arguably most human learning takes place in childhood.

Here, we present a methodology based on DeepMind Lab \citep{beattie2016deepmind} for directly comparing child and agent behavior in simulated exploration tasks, allowing us to precisely test questions about how children explore, how agents explore, and how and why they differ.
Using this methodology, we propose two candidate experiments designed to test key qualitative predictions of different exploration algorithms with respect to what is known about children's exploration behavior in other domains.
Although we are still collecting data in these experiments, we present some preliminary analyses which already raise interesting questions, setting the stage for further research to inspire new approaches to the problem of exploration.

\begin{figure}
    \vspace{-5pt}
    \centering
    \minipage{0.75\textwidth}
        \includegraphics[width=\linewidth]{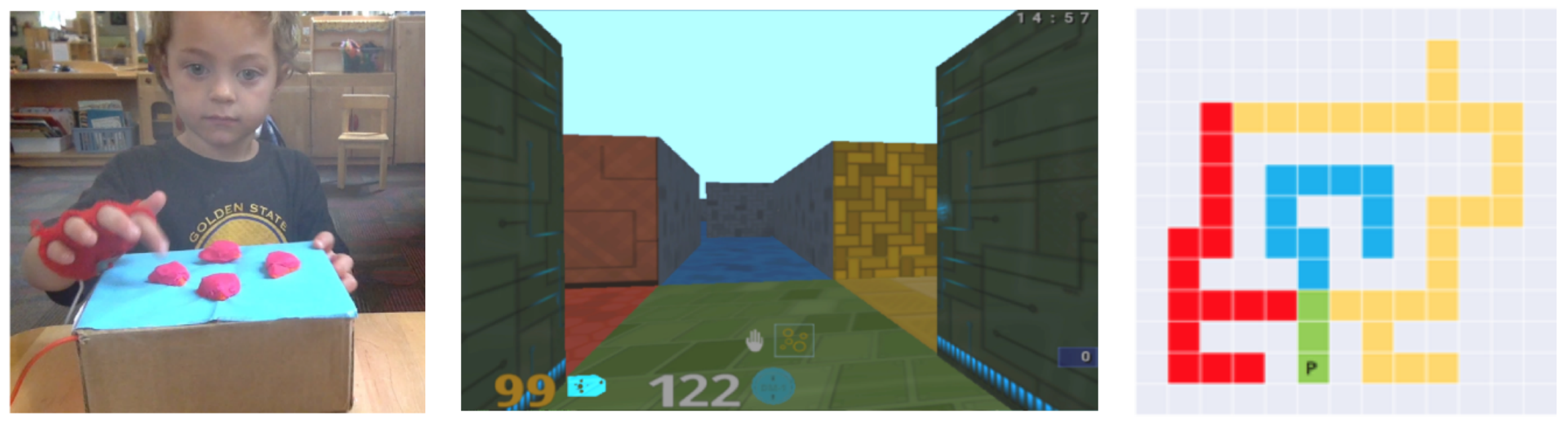}
    \endminipage\hfill
    \caption{(Left) Child using the Arduino-based controller to explore a maze in DeepMind Lab. (Middle) The maze that the child sees on the screen. (Right) Top-down view of maze layout.}
    \label{fig:Kid_mazes}
    \vspace{-15pt}
\end{figure}

\vspace{-2mm}
\section{Directly Comparing Children and Agents}
\label{sec:questions}

We propose using DeepMind Lab \citep{beattie2016deepmind} as a unified environment for training and evaluating both humans and agents.
DeepMind Lab is a learning environment, based on the Quake game engine, that provides a suite of challenging 3D navigation and puzzle-solving tasks for learning agents.
These tasks require physical or spatial navigation capabilities to achieve and are modeled after games that children play themselves. 
In our experimental setup, children are allowed to interact with the DeepMind Lab environment through a custom Arduino-based controller shown in \autoref{fig:Kid_mazes}.
This controller exposes the same four actions that agents would use in this environment (move forward, move back, move left, and turn right).
More technical details about the environment are available in Appendix~\ref{app:environment}.

DeepMind Lab allows us to place agents and humans on more even footing, enabling a more precise exploration of the differences in child and agent behavior.
In particular, we emphasize that the Lab environment is more ecologically valid than other standard RL environments, that it enables more controlled comparisons than are typically seen in the RL literature, and that it provides an avenue for developing new cognitive models.

\textbf{Ecological validity.}
One key reason for gathering data from children and agents in the same environment is that it forces agents to be evaluated in a more ecologically valid setting, compared to grid world-like settings or 2D Atari games \citep{bellemare2013arcade}.
DeepMind Lab combines rich visuals of a simulated world with first-person view, which is much closer to the situated nature of human experience (and which may play an important role in generalization; see \citet{hill2019emergent}).
Furthermore, navigating through the mazes in DeepMind lab is sufficiently interesting such that the human children are captivated by the task.

\textbf{Controlled comparisons.}
Comparing children directly to agents provides a standard baseline for the evaluation of agent behavior, and can assist in identifying areas of promising research in deep RL.
For example, while much of the literature in developmental psychology has focused on free exploration behavior, the majority of work on exploration in artificial intelligence and machine learning has been for goal-seeking domains.
Thus, the quality of an exploration method is typically measured by how much it improves the learning speed and final performance of an agent on a particular task, rather than how well it enables this agent to acquire knowledge and generalize to other tasks.
While recent work in meta-learning \citep[e.g.][]{finn2017model} has begun to expand the metrics that we use beyond single-task reward to transfer efficiency, such papers often lack a strong baseline for human performance and behavior.

\textbf{Cognitive modeling.}
In addition to allowing for an ecologically valid experimental setting, these direct comparisons give strong direction in the development of new cognitive models of behavior, furthering the ``virtuous cycle'' between cognitive science and AI \citep{hassabis2017neuroscience}.
Collecting experimental data from children in the exact environment for which we will test our artificial agents allows us to directly evaluate learned behavior as well as design challenging test-time environments to understand the circumstances when the agent behavior and child behavior diverge.
These divergences have the ability to shed light on issues in both RL and cognitive sciences: How do RL agents react to classical pitfalls for humans?
How do humans react to the classical pitfalls for artificial agents?
Can we create a unified theory? 

\vspace{-2mm}
\section{Illustrative Experiments and Results}
\label{sec:experiments}

Although we are still in the process of collecting and analyzing the data, our preliminary results begin to demonstrate the utility of a direct comparison between children and agents. Both experiments below have been approved by UC Berkeley's IRB.

\vspace{-2mm}
\subsection{Experiment 1: Free versus goal-directed exploration}
\label{sec:exp_1}

Our first experiment is designed to determine if there are differences in the exploration strategies of children who are faced with an unknown environment.
In this experiment, children explored the virtual DeepMind Lab mazes using a custom-built  child-friendly controller (see Appendix~\ref{app:exp1} for full details and maze layouts).
They completed two mazes one after another, each with the same layout.
In the first maze, they were told to explore freely (the ``no-goal'' condition), and in the second maze they were told to search for a ``gummy'' (the ``goal'' condition).
Our initial results suggest that children exhibit a wide range of variability in how much they explore in the no-goal condition, with ``low explorers'' only exploring about 22\% of the maze, ``medium explorers'' exploring about 44\%, and ``high explorers'' exploring up to 71\% of the maze.
We see a  relationship between the level of exploration in the no-goal condition and the steps taken to find the gummy in the goal condition: low explorers take the longest amount of time to reach the goal (95 steps on average), and medium explorers take 89 steps, whereas high explorers take 66 steps on average.

We also find that children's search strategies between the no-goal and goal condition differ.
We compared children's behavior to that of a depth-first search (DFS) agent, which pursues an unexplored path until it reaches a dead-end, at which point it will turn around and explore the last unexplored path it has seen.
More details of this agent, and analysis of the experiment are available in Appendix~\ref{app:exp1}.
We find that in the no-goal condition, children made choices consistent with DFS 89.61\% of the time compared to the goal condition, in which children made choices consistent with DFS 96.04\% of the time ($p = 0.0073$).

In combination, the above two results suggest that during the no-goal condition children build a mental model of the maze, which in ``high explorers'' is necessarily more complete.
During the goal condition, children are able to leverage this mental model to perform more efficient, DFS-like search towards the goal. These preliminary results suggest that we can start to understand the children's behavior in terms of existing algorithms for search and exploration.
In contrast, RL agents are unlikely to exhibit directed, efficient exploration. Most state-of-the-art approaches for guiding exploration in RL agents (Sec.~\ref{sec:intro}) depend on the agent first stumbling upon an interesting area by chance, and then encourage the agent to revisit that area until it is no longer ``interesting.'' In other words, RL agents are retrospective, rather than prospective, explorers.

\vspace{-2mm}
\subsection{Experiment 2: Sparse versus dense rewards}

RL agents typically learn best using dense reward signals.
However, dense rewards make agents less incentivized to explore, and can thus lead to poor generalization.
We are interested in characterizing to what extent dense rewards can impact the exploration behavior in children.
If children are less susceptible to over-fitting to dense rewards, their behavior could shed light on how to design RL algorithms with better generalization.

To test this, we developed a second experiment on children aged 4-6, in which children complete two mazes, each with three phases. In the first phase, the children explore the maze either in a ``no-goal'' condition, where there is no goal; a ``sparse''  condition where a goal exists, but there is no local reward; or a ``dense'' condition where a goal and local rewards leading to the goal are present (see Appendix~\ref{app:exp2} for specific experiment details).
In the second phase, children are asked to find the goal again, which is in the same location as during exploration.
In the final phase, they are asked to find the goal, but the optimal route to the goal is blocked.
We hypothesize that both children and RL agents in the ``dense'' condition will (1) follow the dense-reward path directly to the goal in the first phase, (2) find the goal faster in the second phase (since they can repeat the previous dense-reward path), but (3) will take longer to find the goal in the final phase, compared to those in the ``sparse'' condition, because the previous dense-reward path to the goal is now blocked. Some RL agents in the ``dense'' condition may not find the goal at all in the final phase, if they are unable to switch from exploitation to exploration when they find that the path is blocked.

While we are still collecting data, initial experimental data suggest that children are less likely to explore an area in the dense rewards condition, however, surprisingly, that lack of exploration does not hurt their performance in the final phase.

\vspace{-2mm}
\section{Discussion}

Even the most sophisticated methods for exploration in RL tend to explore only in the service of a specific goal, and are usually driven by error rather than seeking information.
We believe that to truly build intelligent agents, they must do as children do: actively explore their environments, perform experiments, and gather information to weave together into a rich model of the world, which can later be used to rapidly perform new tasks.
Our proposed paradigm using DeepMind Lab to support this endeavor by allowing us to identify the areas where agents and children already act similarly and those in which they do not.
Indeed, our preliminary results already suggest qualitative differences between the exploration behavior of children and agents: for example, we expect that most deep RL agents will not replicate the DFS-like behavior that we observed in children in Experiment 1.

This work only begins to touch on a number of deep questions regarding how children and agents explore.
The two experiments presented here touch on questions of how much children and agents are willing to explore; whether free versus goal-directed exploration strategies differ; and how reward shaping affects exploration.
Yet, our setup allows us to ask so many more, and we have concrete plans to do so.
These include: how easily do agents and children get distracted by irrelevant stimuli or objects in a maze?
To what extent can children and agents remember and integrate information during exploration to aid in future tasks?
How do children react to both positive and negative rewards, and explore mazes safely?
In asking these questions, we will be able to acquire a deeper understanding of the way that children and agents explore novel environments, and how to close the gap between them.

\section{Acknowledgements}

We would like to thank David Szepesvari and Alyosha Efros for helpful discussions and feedback on this work. 

This material is based upon work supported by the 
DARPA – Machine Common Sense grant (HR001119S0005) and the National Science Foundation Graduate Research Fellowship under Grant No. 1752814.

\bibliography{iclr2020_conference}
\bibliographystyle{iclr2020_conference}

\clearpage
\appendix
\section*{Appendix/Supplementary Materials}

\section{DeepMind Lab Environment}
\label{app:environment}

\begin{figure}[h]
    \centering
    \minipage{0.82\textwidth}
        \includegraphics[width=\linewidth]{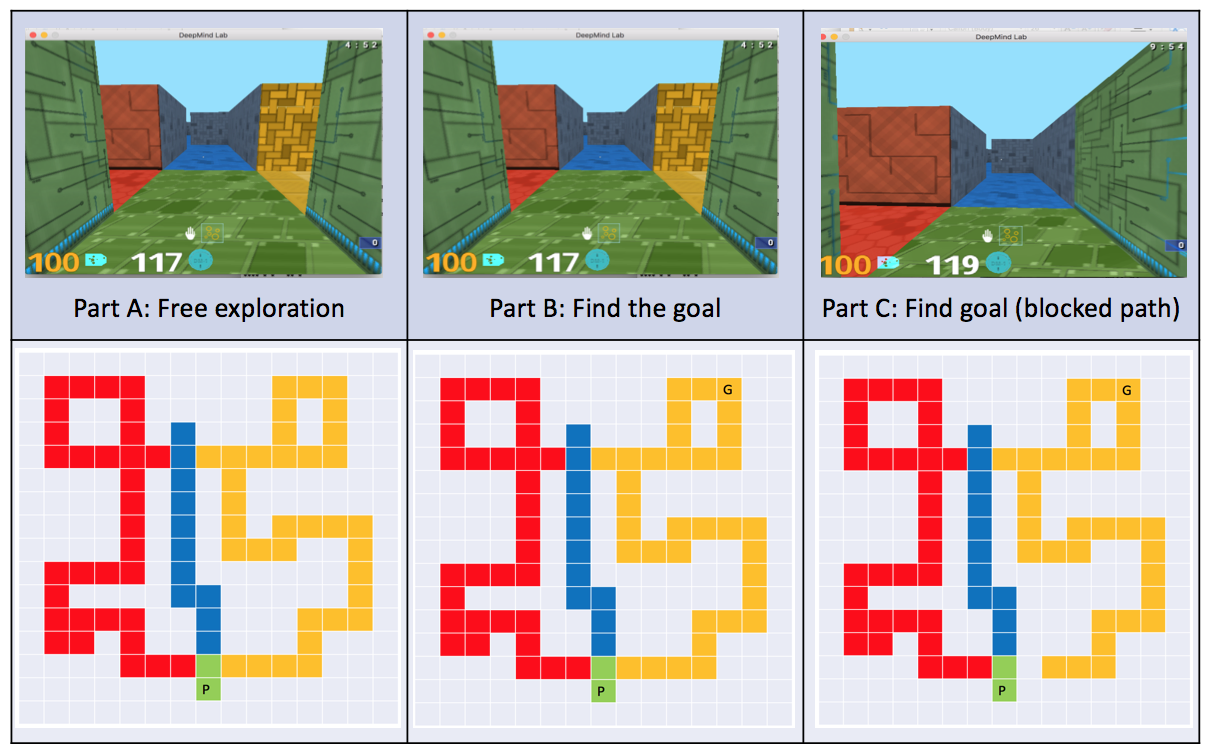}
    \endminipage\hfill
    \caption{(Top) What the child sees when they start each of 3 different parts of the maze game in DeepMind Lab (Bottom) Top-down view of maze layout for each of the 3 parts of the experiment}
    \label{fig:exp1}
\end{figure}

Observations in the DeepMind Lab environment are rendered at 30FPS (close to human perception), and actions cause the avatar in the maze to move either forward or back or to turn left or right. These actions cause the avatar to move 10-15 game units forward/backward in the game space, which is equivalent to about 1/5th of a cell in Figure \ref{fig:exp1}.
As either children or agents can interact with DeepMind lab, in both cases we record the same type of state, action and trajectory information.
Trajectories are discretized by determining which cell the avatar is in (see Figure \ref{fig:exp1}), and recording the new state if it is not equivalent to the cell at the previous time step. These trajectories are then directly compared.

\section{Experimental Details}

\subsection{Experiment 1}
\label{app:exp1}

Our first experiment is designed to determine if there are differences in the exploration strategies of children when faced with an unknown environment. 
In this experiment, children explored the virtual DeepMind Lab mazes using our child-friendly controller.
They completed three mazes one after another, each with the same over-all layout.
In the first maze, they were told to explore freely (the ``no-goal'' condition), and in the second maze they were told to search for a ``gummy'' (the ``goal'' condition).
 and (c) search for the gummy when the most direct path to the goal is blocked.
This ``blocked'' condition design is directly inspired by Tolman mazes \citep{Tollman1946} designed for rats, which demonstrate the ability to re-orient themselves to find a reward in a blocked condition. 
For this experiment, we pre-registered a sample size of 50 children aged 4-5. 

Our initial results suggest that children exhibit a wide range of variability in how much they explore in the no-goal condition, using K-means we break the children into 3 types of explorers (low, medium and high) based on how much they explored in the first ``no goal'' part.  ``Low explorers'' only explored about 22\% of the maze, ``medium explorers'' explored about 44\%, and ``high explorers'' explored up to 71\% of the maze.
Low explorers take about 94.89 steps on average to reach the goal, medium explorers take 89.4 steps and high explorers take about 66.01 steps.
This suggests that naturally exploring more in Part A (without prompts) helps you find the goal in Part B. We do not find any correlations between explorer type and steps taken to reach the goal in Part C in the blocked condition, but plan to explore this further.

\textbf{Comparison to Depth First Search:} Depth first search is a systematic search algorithm which operates by greedily traversing a path until no further traversal can be made. It then backtracks to the most recent branching point which has unexplored branches, and then explores down a new, unexplored branch. 

In addition to the metrics mentioned above, we also compute what we call ``Consistency'' between a child's behavior and the depth first search algorithm. This metric stands in as a proxy for systematic behavior: and attempts to examine if children explore in systematic ways in a maze. Indeed, depth first search is an extremely efficient way to locally explore a maze (unlike breadth first search, which hops around the open list, and would be difficult for a child to replicate). To compute this metric, we begin by discretizing the space of the maze into cells, and computing the trajectory for the child based on the order of cells visited. For a child's trajectory, a step in the child's trajectory is ``consistent'' with the depth first search algorithm if: 
\begin{enumerate}
    \item The child does not visit a state they have previously visited UNLESS there are are no adjacent un-visited neighbors.
    \item if all neighboring states have been visited, the child moves in the direction of the most recent unexplored branch.
\end{enumerate}
Measuring consistency across a child's entire trajectory would lead to large numbers of consistent states (as they traverse down long corridors), making it difficult to measure the actual behavioral differences of the children and the agents. To avoid this confound, we restrict our analyses to ``decision points'' in the maze, that is cells that do not have two neighboring cells. The code for this analysis is made available at \url{https://github.com/CannyLab/ExpExp}. Comparisons to other local search algorithms such as jump-point search are also an interesting avenue for future work. 

\subsection{Experiment 2}
\label{app:exp2}

\begin{figure}
    \centering
    \minipage{0.82\textwidth}
        \includegraphics[width=\linewidth]{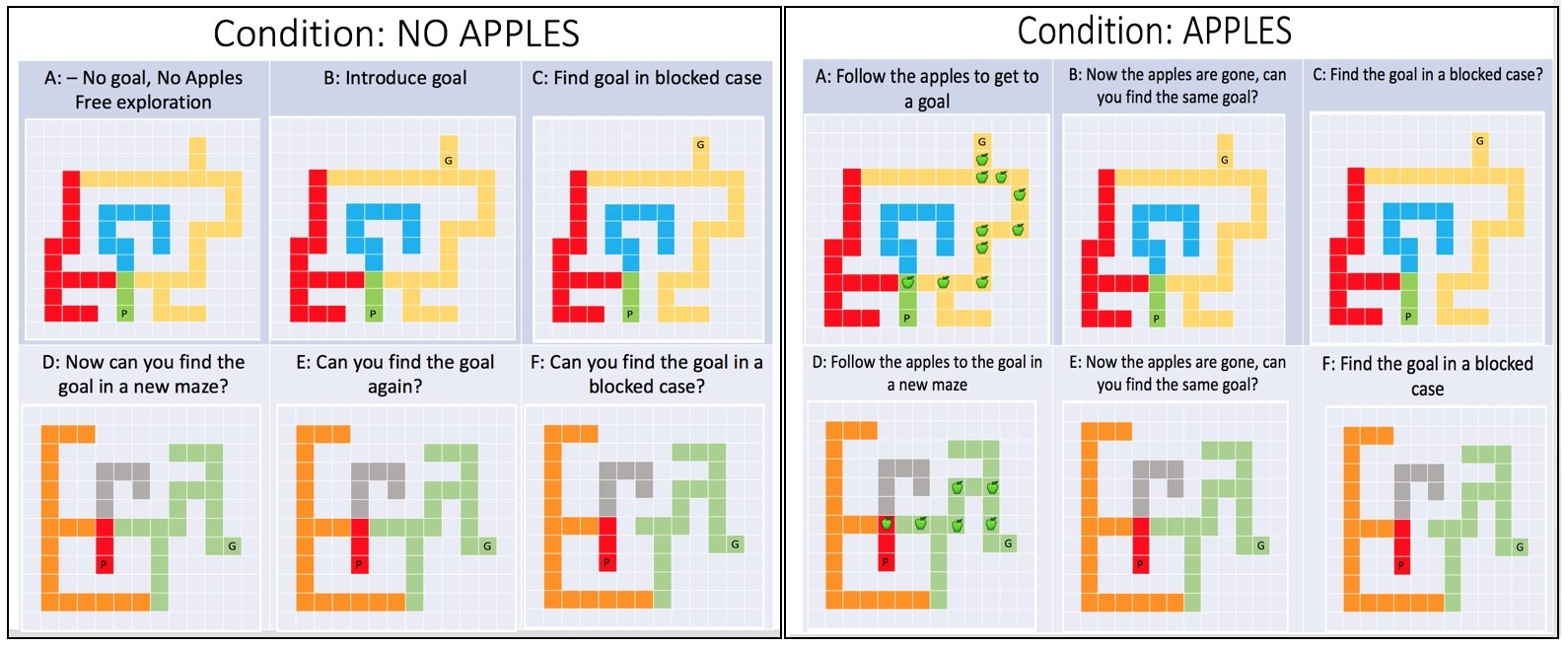}
    \endminipage\hfill
    \caption{(Left) Maze outlines for No apples or ``sparse reward'' condition. (Right) Maze outlines for Apples or ``dense reward'' condition}
    \label{fig:exp2}
\end{figure}

In our second pilot experiment, we pre-registered a sample size of 60 children aged 4-6. Designed to examine the impact of a dense versus sparse reward structure on the exploration patterns of both children and agents.

This task contains 6 parts. Children complete two mazes, each with three phases. There are 2 conditions. The ``dense rewards'' or ``apples'' condition has children following a path of apples in the maze which leads them to the goal. These apples are taken away in the subsequent maze but the goal remains in the same place, children have to find the goal without the aid of the apples. They then have to find the goal in a maze where the main path to the goal is now blocked from the start. In the ``sparse reward'' or ``no apples'' condition, children freely explore a maze that has no goal present. They are then asked to look for a goal in the subsequent maze.  They then have to find the goal in a maze where the main path to the goal is now blocked from the start. For both conditions the same tasks are repeated in a new maze design to test for generalization. Figure 3 outlines what the maze outlines look like. We will measure percent of maze explored for each part, number of steps taken, cells crossed, time to reach goal and percent of maze re-explored in each section. 

We expect that, in line with state-of-the-art RL agents, children in the ``dense'' condition will (1) follow the dense-reward path directly to the goal in the first phase, (2) find the goal faster in the second phase (since they can repeat the previous dense-reward path), but (3) will take longer to find the goal in the final phase, compared to children in the ``sparse'' condition, because the previous dense-reward path to the goal is now blocked.
While we are still collecting data, initial experimental data suggest that children are less likely to explore an area in the dense rewards condition, but, surprisingly, that does not hurt their performance in the final phase.

While we do not know why the lack of exploration does not hurt their performance in the final phase, this shows experiment shows us why its useful to study children, their behavior here is surprising, and certainly an instance where agents behavior differs from children.

\end{document}